\documentclass[10pt, conference]{IEEEtran}
\usepackage[sort&compress,square,numbers]{natbib}
\usepackage{amsmath,amssymb,amsfonts}
\usepackage{algorithmic}
\usepackage{graphicx}
\usepackage{textcomp}
\usepackage{xcolor}
\def\BibTeX{{\rm B\kern-.05em{\sc i\kern-.025em b}\kern-.08em
    T\kern-.1667em\lower.7ex\hbox{E}\kern-.125emX}}

\usepackage{bm}
\usepackage{soul}

\usepackage[linesnumbered,ruled,vlined]{algorithm2e}

\usepackage{adjustbox}
\usepackage{hyperref}
\usepackage{url}
\usepackage{tablefootnote}

\usepackage{caption}
\usepackage{makecell}
\usepackage{multirow}
\begin{document}

\title{Multi-Robot Trajectory Generation via Consensus ADMM: Convex vs. Non-Convex
}

\author{\IEEEauthorblockN{Jushan Chen}
\IEEEauthorblockA{\textit{Dept. of Aerospace Engineering, University of Illinois Urbana-Champaign}\\
jushanc2@illinois.edu}

}

\maketitle

\begin{abstract}
C-ADMM is a well-known distributed optimization framework due to its guaranteed convergence in convex optimization problems. Recently, C-ADMM has been studied in robotics applications such as multi-vehicle target tracking and collaborative manipulation tasks. However, few works have investigated the performance of C-ADMM applied to non-convex problems in robotics applications due to a lack of theoretical guarantees. For this project, we aim to quantitatively explore and examine the convergence behavior of non-convex C-ADMM through the scope of distributed multi-robot trajectory planning. We propose a convex trajectory planning problem by leveraging C-ADMM and Buffered Voronoi Cells (BVCs) to get around the non-convex collision avoidance constraint and compare this convex C-ADMM algorithm to a non-convex C-ADMM baseline with non-convex collision avoidance constraints. We show that the convex C-ADMM algorithm requires 1000 fewer iterations to achieve convergence in a multi-robot waypoint navigation scenario. We also confirm that the non-convex C-ADMM baseline leads to sub-optimal solutions and violation of safety constraints in trajectory generation. 

\end{abstract}

\section{Introduction \& Motivation}
Safe multi-agent trajectory optimization is a challenging problem in many real-world scenarios, such as cooperative inspection and transportation \cite{survey_underwater,aerial_Swarm_survey,non_cooperative_dmpc}. In these settings, each robot must avoid collision to ensure a smooth and safe operation. In addition, each robot must not violate its constraints such as its actuator limits and dynamics. From a trajectory optimization perspective, computational complexity is an important aspect to be accounted for when we have a large-scale multi-robot system, and thus having distributed computation among the robots is highly desired where each agent carries out computation via its local processor while exchanging necessary information with the others.

The consensus alternating direction method of multipliers (C-ADMM) \cite{c_admm} is a popular distributed optimization framework based on the augmented Lagrangian, where each agent exchanges information with its neighbors to iteratively refine its local solution through distributed primal-dual updates. C-ADMM does not require a central node or central processor to coordinate between each agent, and thus each agent optimizes its own local problem in a fully distributed fashion. It is well-established that C-ADMM has guaranteed convergence for convex problems. C-ADMM has been applied in multi-robot applications such as collaborative manipulation, model predictive control, and multi-target tracking \cite{ADMM_manipulation,dmpc_consensus,SOVA_dmpc,Shorinwa2020DistributedMT}. However, these ADMM-based methods would likely suffer from a slow convergence rate when applied to trajectory optimization due to the underlying coupled non-convex collision avoidance constraints between the robots; in addition, for general non-convex problems, C-ADMM converges to a sub-optimal solution. On the other hand, distributed multi-agent trajectory planning methods that rely on Buffered Voronoi Cells (BVCs) \cite{voronoi_schwager,deadlock_resolution1} have demonstrated collision avoidance guarantees, where each robot's local linear inequality constraints representing its BVCs ensure that no collision occurs. The resulting trajectory optimization problem is a convex problem that is computationally easy to solve. However, although BVC-based trajectory planning algorithms are distributed and can lead to guaranteed convergence, the lack of iterative communications in traditional BVC-based trajectory planning algorithms implies that at each time step, the underlying solutions computed by each robot locally must not deviate too much from each other, leading to conservative results.
To address the aforementioned challenges, we are motivated to explore whether enforcing BVCs instead of non-convex collision avoidance constraints in the C-ADMM algorithm will improve the convergence rate and solution quality in multi-robot trajectory optimization. To investigate the convergence rate of C-ADMM applied to non-convex and convex trajectory optimization problems, we propose two variants of C-ADMM. In the first variant, we enforce collision avoidance constraints via BVCs. In the second version, we enforce collision avoidance constraints through a coupled non-convex constraint. By comparing the convergence RATES of these two C-ADMM variants, we aim to verify our hypothesis that C-ADMM with BVCs converges faster than its counterpart with non-convex collision avoidance constraints.
This paper is organized as follows. In Sec. \ref{sec:problem_formulation}, we introduce the multi-agent trajectory optimization problem. In Sec. \ref{sec:challenges}, we describe in detail the challenges our method aims to tackle. In Sec. \ref{sec:simulations}, we present several simulation studies that compare the performance of the convex and non-convex C-ADMM variants. 
\vspace{-8pt}
\section{Problem Formulation}
\label{sec:problem_formulation}
\subsection{Multi-robot Trajectory Optimization}
\label{sec:classic_bvc}
We consider a team of $N$ robots where we want to design control inputs $u^i$ for each robot $i\in[N]$ so that each robot's goal is to reach a reference position $p^i_{ref}$ while avoiding collisions with each other. To ensure collision avoidance, the position of robot $i$ and $j$ should satisfy $||p^i - p^j||_2 \geq 2r_{s}$, $\forall i,j \in [N]$, $i\neq j$, where $r_{s}$ is a safety radius. We represent the collision-free configuration for each robot $i$ using the BVCs \cite{voronoi_schwager} formulated as the following:
\begin{equation}
\overline{\mathcal{V}}^i=\left\{p \in \mathbb{R}^2 \mid\left(p-\frac{p^i+p^j}{2}\right)^{\mathrm{T}} p^{i j}+r_{\mathrm{s}}\left\|p^{i j}\right\| \leq 0, \forall j \neq i\right\}
\end{equation}
where $p^{ij} = p^i - p^j$. We model the dynamics of each robot as a double integrator. We denote the state of each robot at sampling time $t$ consisting of position $p^i$ and velocity $v_i$ as $x^i_{t} = [p^i_{t}, v^i_{t}] \in \mathbb{R}^6$. We denote the input of each robot at $t$ consisting of the acceleration $a^i_{t}$ as $u^i_{t} = [a^i_{t}] \in \mathbb{R}^3$. The dynamics of each robot $i$ is then given as:
\begin{equation}
\label{eqn:dynamics}
    x^i_{t+1} = Ax^i_{t} + Bu^i_{t}
\end{equation}
where $A$ and $B$ are constant state and input matrices with appropriate dimensions. We denote the length of a finite horizon as $T$. We then define our trajectory optimization problem for each robot $i\in[N]$ as a quadratic program given as the following:
\begin{equation}
\
\begin{array}{cl}
\min _{\left\{\boldsymbol{u}^i, \boldsymbol{x}^i\right\}} & L^i\left(\boldsymbol{u}^i, \boldsymbol{x}^i\right) \\
\text { s.t. } &  p^i_{k} \in \overline{\mathcal{V}}^i, \quad k \in [1,\dots,T] ;\\
& x_{k+1}^i=\mathbf{A} x_{k}^i+\mathbf{B} u_{k}^i, \quad k \in [0,\dots,T-1] ; \\
& \left\|u_k^i\right\| \leq a_{\max },  \quad k \in [0,\dots,T-1]; \\
& x^i_{0} = x^i_{init}

\end{array}
\label{eqn:QP_problem}
\end{equation}
where $L^i(\cdot)$ is a quadratic tracking cost function that robot $i$ tries to minimize and $p^i_k$ is its planned position at time $k$. We adopt the following abbreviated notations $u^i = [u^i_0,u^i_1,\dots, u^i_{T-1}]$ and $x^i = [x^i_0, x^i_1, \dots, x^i_{T}]$ to denote agent $i$'s input and state concatenated over a horizon $T$. We further enforce an actuator limit constraint on each robot $i$ at all time steps $k$ such $u^i_k \leq a_{max} \in \mathbb{R}^3$. The optimal control problem in ~\eqref{eqn:QP_problem} can be solved in a distributed model predictive control fashion, where each agent $i$ applies the first element of the optimal control sequence $u^{i*}$ to forward propagate its dynamics, and re-solves the optimization problem defined in ~\eqref{eqn:QP_problem} by updating the initial condition $x^i_{init}$. This process is repeated until convergence.

\subsection{Variant 1: Convex Trajectory Optimization via C-ADMM with BVCs}
\label{sec:variant1}
In C-ADMM, each agent $i$ receives information from its neighbor $j \in \mathcal{N}^i$ and iteratively refines its local solution via a primal and dual update. We first denote the vertical concatenation of the BVC constraint and the actuator limit constraint in Eqn. \ref{eqn:QP_problem} as $g^i(\cdot) \leq 0$, where $g^i(\cdot) \leq 0$ represents the set of linear inequality constraints imposed on agent $i$. We then denote the dynamics constraint in Eqn. \ref{eqn:QP_problem} as $h^i(\cdot) = 0$, which represents the double integrator dynamics of each agent $i$. To formulate the C-ADMM problem, we denote a global optimization variable $\mathrm{\theta}$ as the concatenation of all robots' state trajectory $x^i$ and input trajectory $u^i$ such that $\mathrm{\theta} = [x^1,x^2,\dots,x^N, u^1,u^2,\dots,u^N]$. In C-ADMM we require that each robot $i$ keeps a \textit{local copy} of the global optimization variable $\theta$ denoted as $\mathrm{\theta}^i$. We formulate the C-ADMM problem as the following:
\begin{equation}
\label{eqn:ADMM_problem_convex}
\begin{array}{ll}
\underset{\theta^1, \cdots, \theta^N}{\operatorname{minimize}} & \sum_{i=1}^N L^i\left(\theta^i\right) \\
\text { subject to } & \theta^i=\theta^j ,\quad \forall j \in \mathcal{N}^i \quad i=1, \cdots, N \\
& g^i(\theta^i) \leq 0 \\
& h^i(\theta^i) = 0
\end{array}
\end{equation}
where $L^i(\theta^i)$ refers to the local objective term of agent $i$ and $\mathcal{N}^i$ denotes agent $i$'s neighborhood set. The constraint $\theta^i = \theta^j$ in Eqn.~\eqref{eqn:ADMM_problem_convex} ensures that each robot $i$'s local copy of the decision variable $\mathrm{\theta}^i$ will reach a consensus with its neighbor $\mathrm{\theta}^j$. We denote the dual variable for each robot $i$ as $\lambda^i$ with appropriate dimensions. The C-ADMM algorithm solves the problem in Eqn.~\eqref{eqn:ADMM_problem_convex} through an iterative procedure by first minimizing an augmented Lagrangian with respect to the primal variable $\theta^i$, and then performing a gradient ascent of a dual variable. In addition, each agent $i$ carries out its computation locally, leading to a fully distributed algorithm. For simplicity of exposition, we skip the derivation and give the resulting iterative primal-dual update for each agent $i$ at each iteration $n$ as follows:
\begin{equation}
\begin{aligned}
\label{eqn:primal_dual_update_convex}
\theta^i_{n+1}=\underset{\theta^i}{\operatorname{argmin}} & \left(L^i\left(\theta^i\right)+\lambda^{i T}_{n} \theta^i\right. \\
& \left.+\rho \sum_{j \in \mathcal{N}^i}\left\|\theta^i-\frac{\theta^i_n+\theta^j_n}{2}\right\|_2^2\right) \\
s.t. \quad  & g^i(\theta^i) \leq 0 \\
& h^i(\theta^i) = 0
\end{aligned}
\end{equation}

\begin{equation}
\label{eqn:dual_update_convex}
\lambda^i_{n+1}=\lambda^i_n+\rho \sum_{j \in \mathcal{N}^i}\left(\theta^i_{n+1}-\theta^j_{n+1}\right)
\end{equation}
Intuitively, at each primal update defined in Eqn.~\eqref{eqn:primal_dual_update_convex}, each robot solves a constrained optimization problem based on its local objective function $L^i(\theta^i)$ and the information it received from its neighbors. The dynamics constraint $h^i(\cdot)$ and the concatenation of inequality constraints $g^i(\cdot)$ are enforced in the primal update~\eqref{eqn:primal_dual_update_convex} to ensure feasibility of the trajectory planning problem. At each dual update defined in Eqn.~\eqref{eqn:dual_update_convex}, agent $i$ updates its local dual variable $\lambda^i$ \textit{locally} based on the consensus error $\sum_{j \in \mathcal{N}^i}(\theta^i - \theta^j)$. This iterative procedure can be terminated when the primal residual $||\theta^i_{n+1} - \theta^j_{n+1}||_2$ and the dual residual $||\lambda^i_{n+1}-\lambda^i{n}||_2$ reach a threshold value $\epsilon$, i.e. $||\theta^i_{n+1} - \theta^j_{n+1}||_2 \leq \epsilon^{primal}$ and $||\lambda^i_{n+1}-\lambda^i_{n}||_2 \leq \epsilon^{residual}$. For convex problems, a few dozen iterations usually satisfy the convergence criterion \cite{multi_robot_survey}.

\subsection{Variant 2: Non-convex trajectory optimization via C-ADMM}
\label{sec:variant2}
For a non-convex trajectory optimization problem, it is required that $||p^i_k - p^j_k||_2 \geq r_{min}$ at any time step $k$, for all $j \in \mathcal{N}^i$, where $r_{min}$ is a safety distance between agent $i$ and agent $j$. This constraint is inherently non-convex, and we impose this constraint denoted as $\phi^i(\theta^i) \leq 0$ to each agent's decision variable $\theta^i$. Intuitively, since $\theta^i$ is a local copy of the global optimization variable $\theta$, the collision avoidance constraint $\phi^i(\theta^i)$ implies that from agent $i$'s perspective, the constraint $||p^i_k - p^j_k||_2 \geq r_{min}$ should be active for all time steps $k = [1, \dots, T]$, for all $j \in \mathcal{N}^i$. The non-convex C-ADMM problem is then formulated as follows:
\begin{equation}
\label{eqn:ADMM_problem_nonconvex}
\begin{array}{ll}
\underset{\theta^1, \cdots, \theta^N}{\operatorname{minimize}} & \sum_{i=1}^N L^i\left(\theta^i\right) \\
\text { subject to } & \theta^i=\theta^j ,\quad \forall j \in \mathcal{N}^i \quad i=1, \cdots, N \\
& g^i(\theta^i) \leq 0 \\
& h^i(\theta^i) = 0 \\
& \phi^i(\theta^i) \leq 0\\
\end{array}
\end{equation}
where $g^i(\theta^i)$ now only denotes the actuator limit constraints. At each iteration $n$, agent $i$ executes the following primal-dual update:
\begin{equation}
\begin{aligned}
\label{eqn:primal_update_nonconvex}
\theta^i_{n+1}=\underset{\theta^i}{\operatorname{argmin}} & \left(L^i\left(\theta^i\right)+\lambda^{i T}_{n} \theta^i\right. \\
& \left.+\rho \sum_{j \in \mathcal{N}^i}\left\|\theta^i-\frac{\theta^i_n+\theta^j_n}{2}\right\|_2^2\right) \\
s.t. \quad  & g^i(\theta^i) \leq 0 \\
& h^i(\theta^i) = 0\\
& \phi^i(\theta^i) \leq 0
\end{aligned}
\end{equation}

\begin{equation}
\label{eqn:dual_update_nonconvex}
\lambda^i_{n+1}=\lambda^i_n+\rho \sum_{j \in \mathcal{N}^i}\left(\theta^i_{n+1}-\theta^j_{n+1}\right)
\end{equation}
Non-convexity thus arises from the primal update defined in~\eqref{eqn:primal_update_nonconvex} due to the non-convex constraint function $\phi^i(\cdot)$. The dual update~\eqref{eqn:primal_update_nonconvex} remains the same as in Eqn.~\eqref{eqn:dual_update_convex}.

\subsection{C-ADMM in receding horizon control}
In Sec. \ref{sec:variant1} and Sec. \ref{sec:variant2}, we presented two variants of C-ADMM algorithms for trajectory optimization, where the former is convex and the latter is non-convex. To formulate C-ADMM in a receding-horizon fashion, we embed the algorithms presented in Eqn.~\eqref{eqn:primal_dual_update_convex},~\eqref{eqn:dual_update_convex} or Eqn.~\eqref{eqn:primal_update_nonconvex},~\eqref{eqn:dual_update_nonconvex} in a receding-horizon loop. As a result, we have an inner loop and an outer loop. In the inner loop, we execute the C-ADMM algorithms. In the outer loop, we run a distributed receding-horizon control update, where each agent $i$ forward propagates its state $x^i$. It is noted that although each agent $i$ keeps a local copy of the decision variable $\theta^i$ which contains the states and control inputs of its neighbors and itself, agent $i$ will only apply the control input relevant to itself and discard any other information. We summarize our C-ADMM algorithm embedded in a receding-horizon loop in Algorithm. \ref{algo:distributed-rec-horizon-control}:
\begin{algorithm}
\SetAlgoNlRelativeSize{0}
\SetAlgoNlRelativeSize{-1}
\SetAlgoNlRelativeSize{-2}

\KwData{System dynamics $x^i_{t+1} = Ax^{i}_{t} + B u^{i}_{t}$, Initial state $x_0$, reference state $x_{ref}$, prediction horizon $T$}
\KwResult{Optimal control sequence $u^i$}
\While{Not converged}{
\For{each sampling time $t$}{
    \ForEach{agent $i \in [N]$}{
        Run C-ADMM updates ~\eqref{eqn:primal_dual_update_convex},~\eqref{eqn:dual_update_convex} or~\eqref{eqn:primal_update_nonconvex},~\eqref{eqn:dual_update_nonconvex} until convergence\;
    }
    \ForEach{agent $i$}{
        Extract optimal local control sequence $u^{i*}_{t:t+T-1}$\;
        Apply the first control input $u^{i*}_t$ to forward propaget its dynamics to $x^i_{t+1}$ \;
    }
}
}
\caption{C-ADMM Receding-Horizon Control}
\label{algo:distributed-rec-horizon-control}
\end{algorithm}

\section{Challenges}
\label{sec:challenges}
While C-ADMM has gained a lot of popularity for distributed optimization applications, applying it directly to multi-robot trajectory planning with coupled collision avoidance constraints poses the following key challenges:
\begin{itemize}
    \item Slow convergence with non-convex constraints: The coupled inter-robot collision avoidance constraints are highly non-convex, which might slow down the convergence of C-ADMM. The exact impact of nonconvexity on the convergence behavior is currently lacking in the trajectory planning literature and it needs further examination.
    \item Sub-optimal solutions: For general non-convex trajectory optimization problems might to a sub-optimal solution. In this project, we tackle the challenge of quantitatively and qualitatively analyzing how sub-optimal the non-convex C-ADMM algorithm is compared to our convex C-ADMM.
\end{itemize}
We hypothesize that our method will tackle these challenges by ensuring collision avoidance via BVCs rather than the conventional non-convex collision avoidance constraints. We 

\section{Simulation studies}
\label{sec:simulations}
We perform several simulation studies by applying the two C-ADMM variants discussed in Sec. \ref{sec:variant1} and Sec. \ref{sec:variant2} to multi-robot waypoint transition scenarios and compare the results both quantitatively and qualitatively. In Sec. \ref{sec:trajectory_sim}, we present a 5-robot waypoint transition scenario solved by both variants and analyze their respective convergence rates. In Sec. \ref{sec:convergence_analysis}, we present a Monte Carlo analysis to quantitatively examine the convergence rates by varying the number of agents across multiple simulation trials.

\begin{figure*}[ht]
    \centering
    \includegraphics[width=0.8\textwidth,height = 13cm]{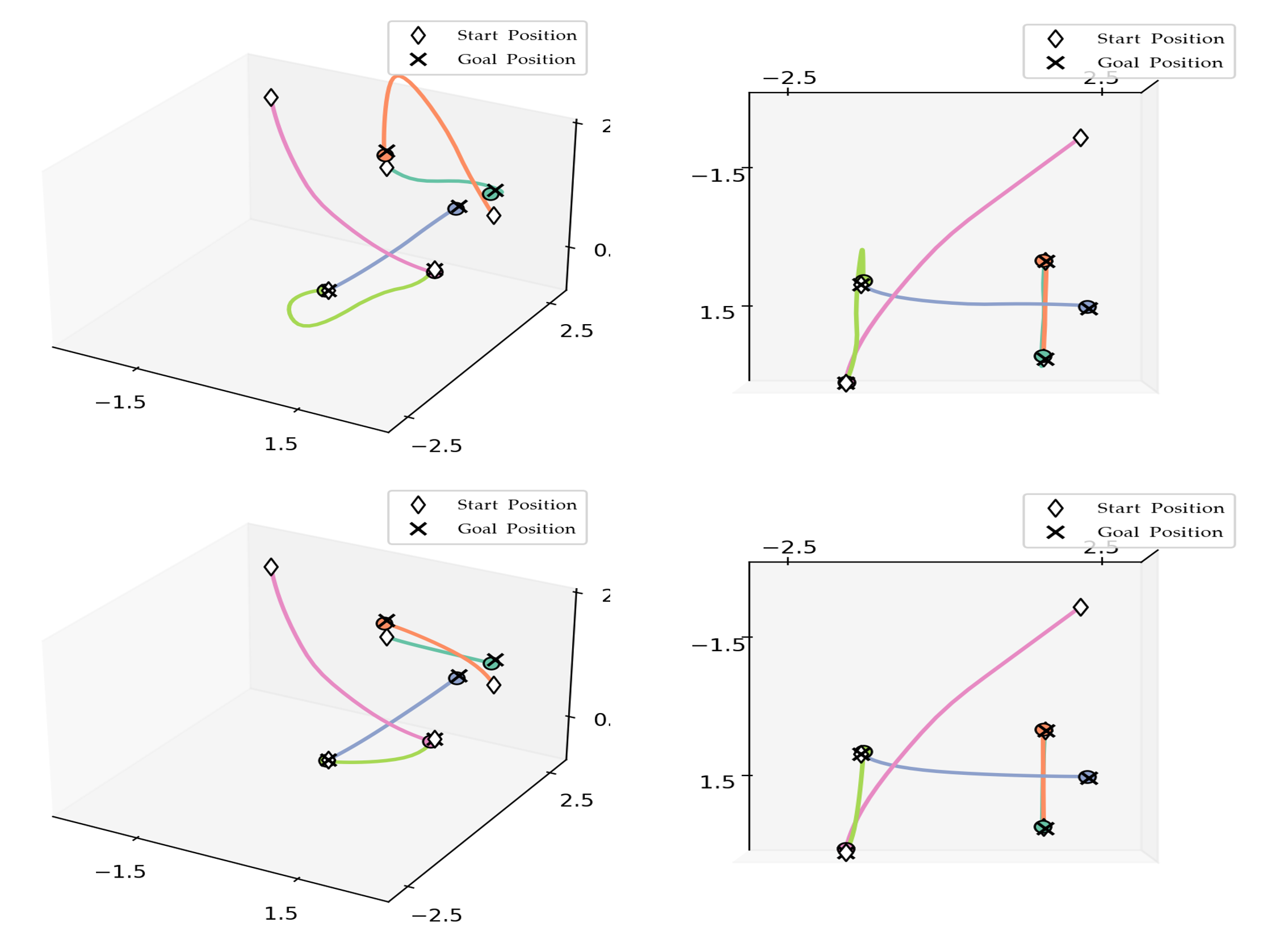}
    \caption{Trajectory comparison of our convex C-ADMM and the non-convex C-ADMM baseline on a 5-drone navigation scenario. From left to right, we show the side view of the trajectory and the bird's eye view of the trajectory.}
    \label{fig:5_drone_Experiment}
\end{figure*}
\subsection{Multi-agent waypoint transition}
\label{sec:trajectory_sim}
\begin{figure}[h]
    \centering
    \includegraphics[height = 12cm, width = 0.5\textwidth]{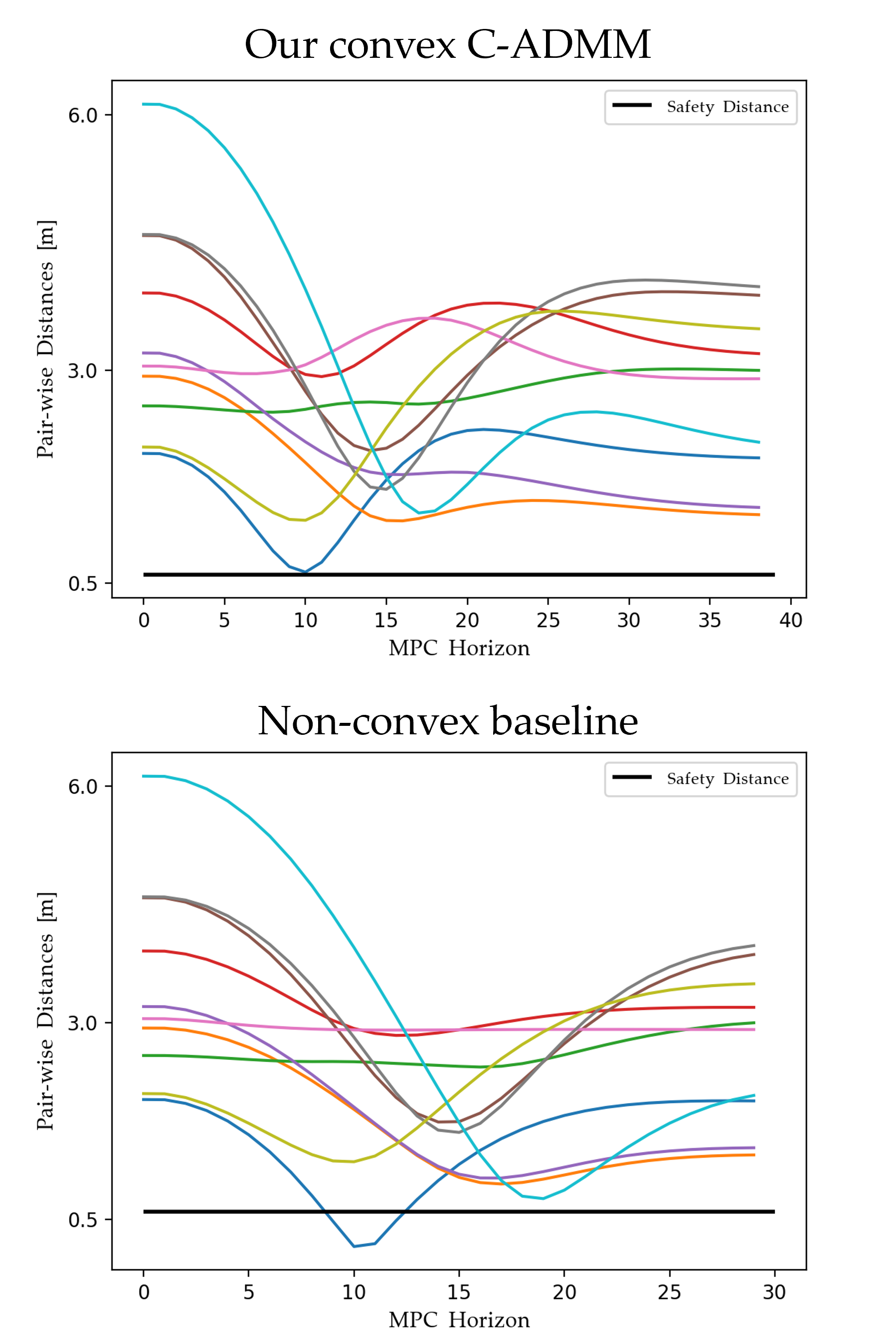}
    \caption{Pairwise distances between all robots in the 5-robot waypoint navigation scenario}
    \label{fig:5_Drone_pairwise_distances}
\end{figure}
We show the predicted trajectories for a 5-robot waypoint navigation scenario in Fig. \ref{fig:5_drone_Experiment}. In this simulation scenario, we randomly generated initial and final conditions in a confined 3D space of dimension 3.5 $m$ $\times$ 3.5 $m$ $\times$ 2.5 $m$ in $\mathbb{R}^3$. We used a prediction horizon of $T = 10$. We did not impose a position constraint on each robot, and therefore the trajectory of each robot was allowed to exceed the 3D space defined earlier. Each agent was modeled with double integrator dynamics~\eqref{eqn:dynamics} with a discretization time step of 0.1 $s$. For the safety constraint in the form of either the BVCs~\eqref{eqn:QP_problem} or the conventional non-convex collision avoidance constraint $||p^i_k - p^j_k||_2 \geq r_{min}$, we used a safety distance of $r_{min} = 0.3$ $m$. For both the convex and non-convex variants, we ran the Algorithm. \ref{algo:distributed-rec-horizon-control} until the sum of local objective terms across all agents $\sum_{i}^{N} L^i(\theta^i)$ could no longer decrease. Fig. \ref{fig:5_drone_Experiment} compares the simulated trajectories of our convex C-ADMM method and the non-convex C-ADMM baseline applied to the 5-robot waypoint transition scenario.

By inspecting Fig. \ref{fig:5_drone_Experiment} qualitatively, we observe that both our convex C-ADMM algorithm and the non-convex C-ADMM baseline achieved convergence, i.e., all robots reached their respective goal positions. We then examine the inter-robot distances between each pair of robots at all time steps shown in Fig. \ref{fig:5_Drone_pairwise_distances} to examine any constraint violations. We observe that the non-convex baseline violates the safety constraint at the 10th MPC horizon, whereas our convex C-ADMM method satisfies the safety constraint at all times. In particular, the inter-robot distances at the 10th MPC iteration of our convex C-ADMM algorithm are lower-bounded by the safety threshold distance. Overall, we find that the non-convex C-ADMM baseline yielded a sub-optimal solution compared to our convex C-ADMM algorithm.
\begin{figure}[h]
    \centering
    \includegraphics[width = 0.5\textwidth, height=12cm]{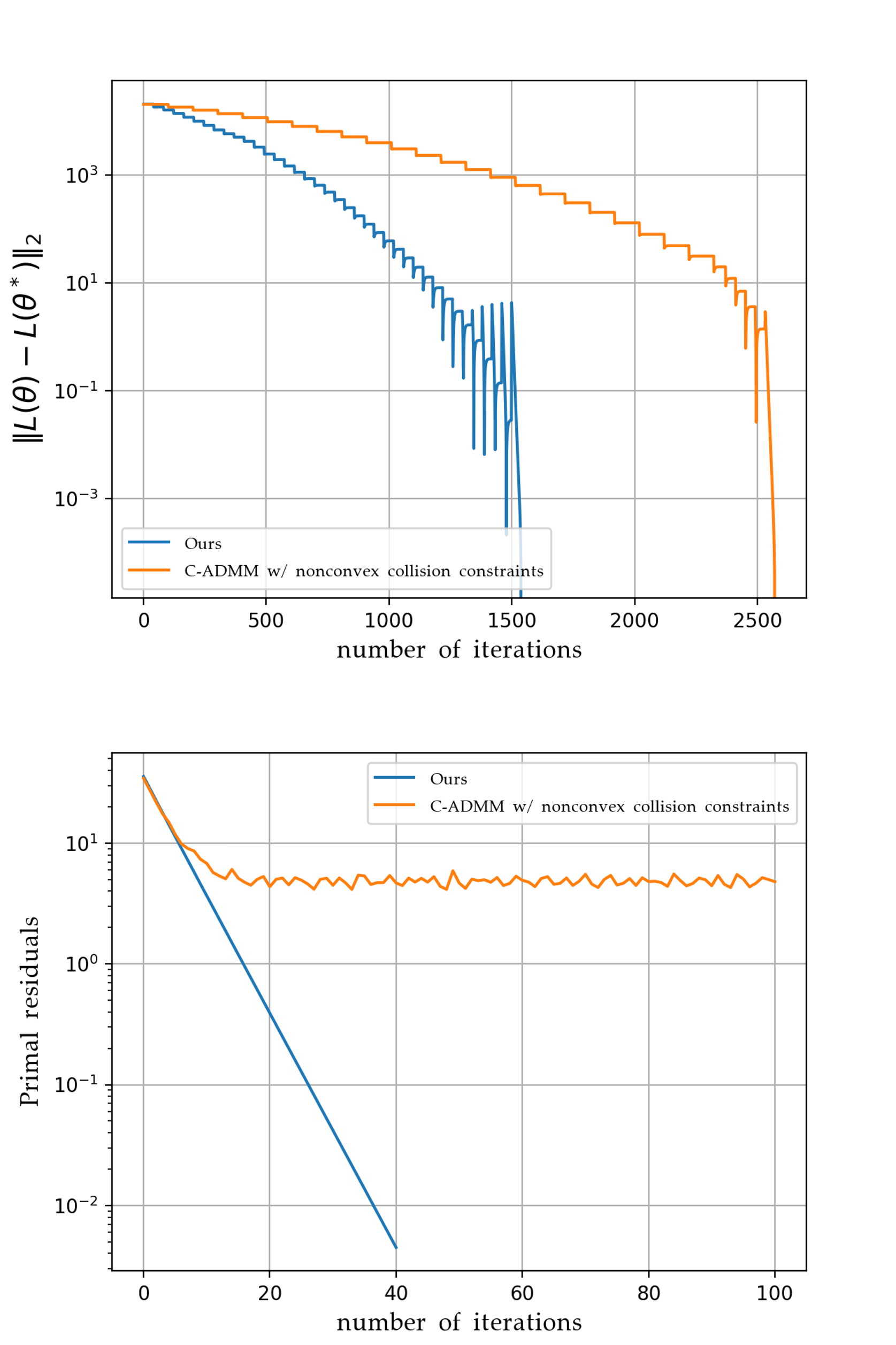}
    \caption{From top to bottom, we show the global objective descent and primal residual descent for both our convex C-ADMM method and the non-convex baseline}
    \label{fig:5_drone_convergence_analysis}
\end{figure}
\subsection{Convergence Analysis}
\label{sec:convergence_analysis}
We then present a convergence analysis in terms of the global objective terms, $L(\theta)$, and the primal residual, $||\theta^i - \theta^j||_2$, $j \in \mathcal{N}^i$ in Fig. \ref{fig:5_drone_convergence_analysis}. We observe that for the 5-robot waypoint navigation scenario, our convex C-ADMM method requires 1000 fewer iterations to reach objective convergence, i.e., a sufficiently small objective value. In addition, we find that our convex C-ADMM method converges to a sufficiently small primal residual value linearly. In contrast, the primal residual for the non-convex C-ADMM baseline fails to converge to a reasonably small primal residual and keeps oscillating around a constant value after a few iterations, which contributes to the sub-optimal solution in the resulting trajectories of the non-convex baseline shown in Sec. \ref{sec:trajectory_sim}. Intuitively, this lack of primal residual convergence indicates that the agents never reached a reasonably good consensus while communicating with each other, which significantly degrades the quality of the solution in practice.

\subsection{Monte Carlo Anlysis}
To quantitatively compare the convergence rates more comprehensively, we present a Monte Carlo simulation by varying the number of robots from 3 agents to 5 agents. We run 40 trials for each number of agents with randomized initial and final conditions for each trial. We record the average number of iterations required until convergence as shown in Table. \ref{table:3_Agent} and Table. \ref{table:5_Agent}.

\begin{table}[h]
\centering
\begin{tabular}{|c|c|}
\hline
 & Avg. number of iterations \\ \hline
Convex C-ADMM (ours) & 701 \\ \hline
\multicolumn{1}{|l|}{Nonconvex C-ADMM} & 1713 \\ \hline
\end{tabular}
\caption{Iterations until convergence for 3-agent waypoint navigation}
\label{table:3_Agent}
\end{table}

\begin{table}[h]
\centering
\begin{tabular}{|c|c|}
\hline
 & Avg. number of iterations \\ \hline
Convex C-ADMM (ours) & 1673 \\ \hline
\multicolumn{1}{|l|}{Nonconvex C-ADMM} & 4984 \\ \hline
\end{tabular}
\caption{Iterations until convergence for 5-agent waypoint navigation}
\label{table:5_Agent}
\end{table}
We observe our convex C-ADMM algorithm consistently requires a much smaller number of iterations until convergence, leading to a faster convergence rate than the non-convex baseline.

\section{Conclusion}
\textbf{Summary}. In this project, we have quantitatively and qualitatively examined the impact of non-convexity in solving multi-robot trajectory optimization problems using C-ADMM from the perspective of solution quality and convergence rates. We conclude that our convex C-ADMM leveraging BVCs leads to a significantly faster convergence rate and more optimal trajectories than the non-convex C-ADMM baseline.\footnote{Code repository at \href{consensus_ADMM}{\texttt{https://github.com/RandyChen233/consensus\_ADMM}}}.
\par
\textbf{Limitations}. We considered a synchronous communication strategy and time-invariant dynamics for the robots in the scope of this project. In the future, we would like to further examine the impact of non-convexity on the convergence of C-ADMM in the presence of time-varying dynamics and asynchronous communication strategies.

\newpage
\bibliographystyle{IEEEtranN}
\bibliography{references}

\end{document}